\documentclass[11pt,a4paper]{article}

\usepackage[utf8]{inputenc}
\usepackage[T1]{fontenc}
\usepackage{graphicx}
\usepackage{amsmath, amssymb}
\usepackage{tabularx}  
\usepackage{booktabs}  
\usepackage{hyperref}
\usepackage{float}
\usepackage{caption}
\usepackage{subcaption}
\usepackage{geometry}
\title{Spatiotemporal Analysis of Forest Machine Operations Using 3D Video Classification}
\author{
Maciej Wielgosz\thanks{Corresponding author with respect to Machine Learning part: \texttt{maciej.wielgosz@nibio.no}}, \\
Simon Berg\thanks{Corresponding author with respect to Forestry part: \texttt{simon.berg@nibio.no}}, \\
Heikki Korpunen, \\
Stephan Hoffmann \\
Norwegian Institute of Bioeconomy Research (NIBIO),\\ Høgskoleveien 8, 1433 Ås
}

\begin{document}

\maketitle

\begin{abstract}
This paper presents a deep learning-based framework for classifying forestry operations from dashcam video footage. Focusing on four key work elements—crane\_out, cutting\_and\_to\_processing, driving, and processing—the approach employs a 3D ResNet-50 architecture implemented with PyTorchVideo. Trained on a manually annotated dataset of field recordings, the model achieves strong performance, with a validation F1 score of 0.88 and precision of 0.90. These results underscore the effectiveness of spatiotemporal convolutional networks for capturing both motion patterns and appearance in real-world forestry environments.

The system integrates standard preprocessing and augmentation techniques to improve generalization, but overfitting is evident, highlighting the need for more training data and better class balance. Despite these challenges, the method demonstrates clear potential for reducing the manual workload associated with traditional time studies, offering a scalable solution for operational monitoring and efficiency analysis in forestry.

This work contributes to the growing application of AI in natural resource management and sets the foundation for future systems capable of real-time activity recognition in forest machinery. Planned improvements include dataset expansion, enhanced regularization, and deployment trials on embedded systems for in-field use.
\end{abstract}


\section{Introduction}\label{sec:intro}  
Time studies have long been a cornerstone of improving efficiency in forest operations, helping to optimize working techniques and machinery usage for better productivity and cost-effectiveness. Traditionally, such studies were conducted with stopwatches and paper logs, often requiring two people to time and record  tasks\cite{mcdonald1999timing}. By the late 20th century, specialized field computers and handheld data loggers began replacing manual timing, streamlining data collection in the field\cite{mcdonald1999timing}. The introduction of video recording technology brought a further enhancement to time studies: video cameras (including modern action cams and dashcams) allow detailed capture of work activities, providing a rich visual record of each work element\cite{mcdonald1999timing}. 

Researchers can replay footage to scrutinize short, specific tasks, which is invaluable for identifying bottlenecks and improving work methods. However, despite these advancements in data collection, the analysis of recorded footage remains labor-intensive. Manually reviewing and annotating videos is time-consuming -- studies often report needing to spend at least three times the video's duration to thoroughly analyze and segment the work elements. This manual approach is not only slow but can also be tedious and error-prone\cite{mcdonald1999timing}, especially as it relies on the subjective judgment of the analyst.

The growing availability of affordable, high-quality action cameras and dashcams in forestry has essentially shifted the bottleneck from data \textit{collection} to data \textit{analysis}. This shift has motivated the exploration of automated methods to assist or replace the human analyst in processing video data. Over the past decade, advances in artificial intelligence -- particularly in deep learning for computer vision -- have enabled the automatic detection and recognition of human activities and machine operations in video footage\cite{bang2019computer}. Harnessing these AI techniques in forestry could dramatically reduce the time required to extract meaningful information from field videos. Instead of a researcher watching hours of footage to identify when a harvester is cutting, processing, moving, or idling, a trained model could flag these work elements automatically. With this goal in mind, we undertook a pilot study to evaluate the feasibility of deep learning-based video analysis for forestry operations.

\section{Related Work}\label{sec:related}
Existing literature offers a spectrum of approaches for automated video analysis in forestry and related domains, ranging from classical machine learning techniques to state-of-the-art deep learning models. In this section, we first discuss classical approaches that rely on hand-crafted features and conventional classifiers, then review modern deep learning methods (CNNs, 3D CNNs, and RNNs) that have been applied to recognize activities in videos, including those pertinent to forest operations.

\subsection{Classical Machine Learning Approaches}
Before the deep learning era, researchers attempted to analyze images and videos using manually designed features fed into traditional classifiers. In the context of action or event recognition, this often meant extracting visual descriptors (e.g., color histograms, texture metrics, edge or shape features, optical flow for motion) and then training algorithms like support vector machines (SVMs), decision trees, or random forests to classify different states or activities. For instance, in the domain of environmental monitoring, Zhao et al. (2011) developed an SVM-based system to automatically detect forest fires in video frames by combining static color/texture cues with dynamic motion features\cite{zhao2011svm}. Their approach used an initial segmentation of potential flame regions, followed by an SVM classifier that learned to distinguish actual fire from confounding objects (like sun glare or moving red objects) based on the temporal variation of the candidate regions\cite{zhao2011svm}. This is an example where a carefully crafted feature pipeline (color, flicker frequency, shape irregularity, etc.) was paired with a classical classifier to address a specific forestry-related video task. Similarly, rule-based decision trees have been used in various contexts to classify phases of work or detect events, especially when a few descriptive variables (e.g., equipment sensor readings or simple image cues) could be thresholded to identify states.

Early efforts in automated time studies for logging machines leveraged sensor data rather than video -- for example, McDonald and Rummer (1999) integrated GPS tracking and mechanical switch sensors on timber harvesters to log different operation phases\cite{mcdonald1999timing}. While effective for gross time measurements (e.g., distinguishing moving vs. cutting), such approaches struggled to reliably discern finer work elements and required significant setup for each piece of equipment. Classical approaches laid important groundwork but also revealed limitations. They generally required extensive feature engineering -- experts had to hypothesize which visual patterns or sensor signals would characterize each class of interest and design algorithms to extract those patterns. This process can be cumbersome and may not generalize well to new scenarios. Moreover, traditional models like SVM or decision trees have trouble handling the full variability and complexity of raw video data. They often rely on simplifying assumptions or require the video to be pre-segmented into clips of single activities. In practice, factors such as changing illumination, background movement (e.g., swaying foliage), and the subtlety of certain operator actions can confound these systems. As a result, the accuracy of early machine learning systems in complex forest environments was limited, and they remained computationally expensive because features had to be extracted from every frame and then classified frame-by-frame\cite{simonyan2014two}. This high computational cost, combined with the potential for missing context between frames, meant that classical methods could only partially automate video analysis. These shortcomings in scalability and robustness motivated researchers to turn toward deep learning methods, which can automatically learn rich feature representations from the raw data.

\subsection{Deep Learning Models for Video Analysis in Forestry}
The advent of deep learning brought significant breakthroughs in image and video recognition, and forestry researchers have begun leveraging these advances. Convolutional Neural Networks (CNNs) excel at extracting hierarchical features from imagery, and early applications in forestry showed their promise on static images. For example, Onishi and Ise employed a deep CNN to automatically classify tree species from aerial images taken by a UAV\cite{onishi2018automatic}. In their 2018 study, a standard RGB camera on a drone was used to capture canopy images, and after segmenting individual tree crowns, a CNN was able to distinguish seven tree species with about 89\% accuracy\cite{onishi2018automatic} -- notably high given that only consumer-grade imagery was used. In another study, Faizal demonstrated that deep learning could even be applied to microscopic-level details like bark texture: using a fine-tuned ResNet-101 CNN, 50 tree species were recognized from bark images with over 94\% accuracy\cite{faizal2022automated}. These works (Onishi et al. and Faizal) highlight how image-based classification tasks in forestry have benefited from CNNs, achieving accuracy levels impossible with earlier hand-crafted approaches. 

Moving from images to full video sequences, deep learning provides two main strategies to incorporate the time dimension: spatiotemporal CNNs and recurrent neural networks. The first strategy extends 2D CNNs into the time domain, yielding 3D CNNs that perform convolution across both space and time. Such networks (pioneered by models like C3D~\cite{tran2015learning} and later refined by others~\cite{tran2018closer}) can directly learn motion patterns from sequences of frames. They have been applied to generic action recognition tasks with great success, and are well-suited to tasks like recognizing machinery motions or worker actions in forestry videos. The second strategy couples CNNs (for spatial feature extraction per frame) with Recurrent Neural Networks (RNNs) such as Long Short-Term Memory (LSTM) units that learn temporal dynamics. In a CNN+LSTM architecture, the CNN might produce a descriptor for each video frame (or short clip), which the LSTM then processes as a time series to classify the overall action. Such hybrid models have proven effective in many domains for capturing both appearance and motion; for example, an integrated 3D CNN-LSTM framework was recently shown to achieve state-of-the-art performance (over 95\% accuracy on benchmark datasets) in complex action recognition tasks~\cite{xu2019multi,kalfaoglu2020late, Sensors2022,FutureInternet2019}.

The forestry domain is beginning to see analogous implementations: one can envision an LSTM that learns the sequence ``crane reaches tree -> saw cuts -> log processes -> crane swings away'' as a pattern distinguishing a cutting cycle from other activities. Several studies in related fields illustrate how deep video models can be applied to outdoor and operational footage. Park and Bang, for instance, utilized a fully convolutional network to analyze dashcam video from vehicles, segmenting out road regions frame-by-frame for autonomous driving purposes\cite{bang2019computer}. Their 2019 approach shows that with enough training data, CNN-based models can robustly interpret video streams from a moving camera -- a scenario not unlike a forest machine-mounted camera capturing its surroundings. In the context of forest operations, we are now seeing the first attempts to use deep learning for work phase detection. Our pilot study contributes to this line of research by applying a modern video classification pipeline (based on PyTorch \cite{falcon2019pytorch} and PyTorchVideo \cite{pytorchvideo2021} libraries) to automatically detect logger work elements in dashcam footage.

This builds on the aforementioned advances: the model we developed uses a ResNet-based 3D CNN to simultaneously encode visual appearance and motion from consecutive frames, and it was trained to recognize classes such as crane-out, processing, and driving. While detailed results are presented in later sections, the takeaway from prior work is clear: deep learning approaches outperform classical methods in handling the variability of forestry videos, thanks to their ability to learn features directly from data. Recent surveys of AI in forestry underscore this trend, noting successful applications of computer vision ranging from tree species mapping to autonomous forest inventory assessments\cite{onishi2018automatic,faizal2022automated}. Our work extends these successes into the realm of time-and-motion analysis, demonstrating how video classification networks can assist in automating time studies. 

\section{Dataset}

Approximately two hours of video footage were recorded from a harvester operating under good field conditions, using a Nextbase 622GW dashcam at 1920\(\times\)1080 resolution and 30\,fps. The footage was manually annotated to label discrete work elements. These annotated clips were then split into training and validation subsets to support model development and evaluation. This pilot dataset served as the foundation for testing a deep learning model capable of recognizing work elements directly from raw video data.

The discrete work elements and their corresponding clip counts are summarized in Table~\ref{tab:dataset_summary}.

\begin{table}[H]
    \centering
    \caption{Summary of dataset classes with definitions and clip counts}
    \label{tab:dataset_summary}
    \begin{tabularx}{\textwidth}{l X r}
        \toprule
        \textbf{Class} & \textbf{Definition} & \textbf{Clips} \\
        \midrule
        crane\_out & Starts when the crane begins to move towards a tree and ends the first time feed rollers touch the tree. & 80 \\
        cutting\_and\_to\_processing & Starts when \textbf{crane\_out} ends and ends the first time feed rollers move. & 78 \\
        processing & Starts when \textbf{cutting\_and\_to\_processing} ends and ends when the last log is bucked. & 72 \\
        driving & When wheels are moving and no other work element is ongoing. & 78 \\
        non\_productive & When the machine is not moving. & 4 \\
        other\_crane\_movement & Any other crane movement not associated with \textbf{crane\_out}. & 37 \\
        \bottomrule
    \end{tabularx}
\end{table}

This distribution demonstrates a reasonably balanced dataset across the major classes, with the exception of \textbf{non\_productive}, which is underrepresented and may introduce challenges in achieving consistent model performance across all categories.

\begin{figure}[H]
    \centering
    \includegraphics[width=0.75\linewidth]{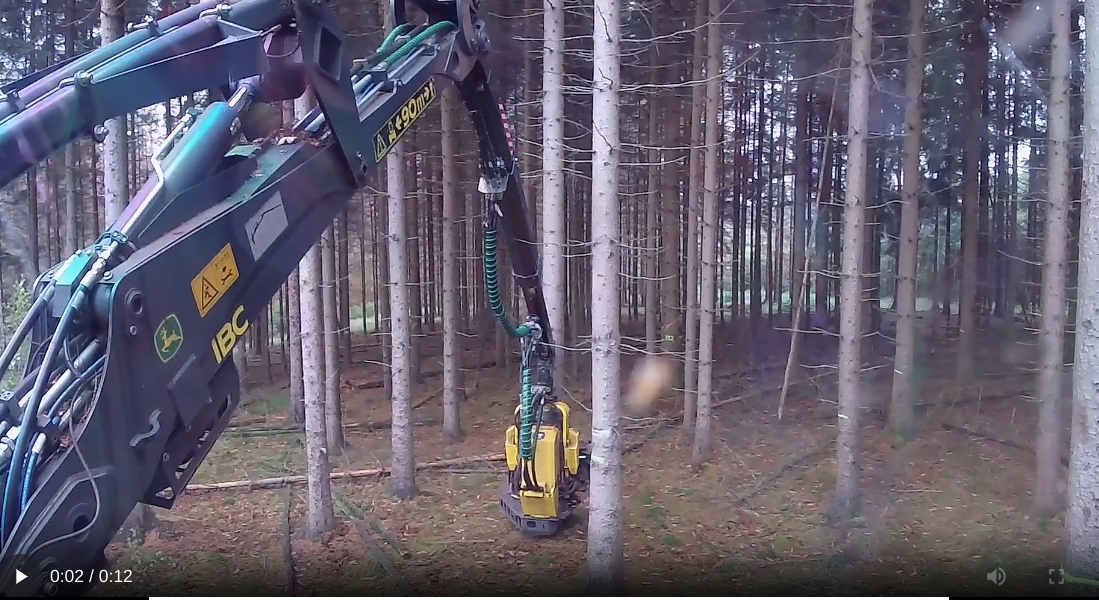}
    \caption{A sample frame from the video showing crane\_out activity}
    \label{fig:crane-out}
\end{figure}

\vspace{0.5em}

For the purposes of model training and evaluation, a subset of four primary classes was used: \textbf{crane\_out}, \textbf{cutting\_and\_to\_processing}, \textbf{driving}, and \textbf{processing}. These classes were chosen based on their frequency and operational significance in field workflows.

\begin{figure}[H]
    \centering
    \includegraphics[width=\linewidth]{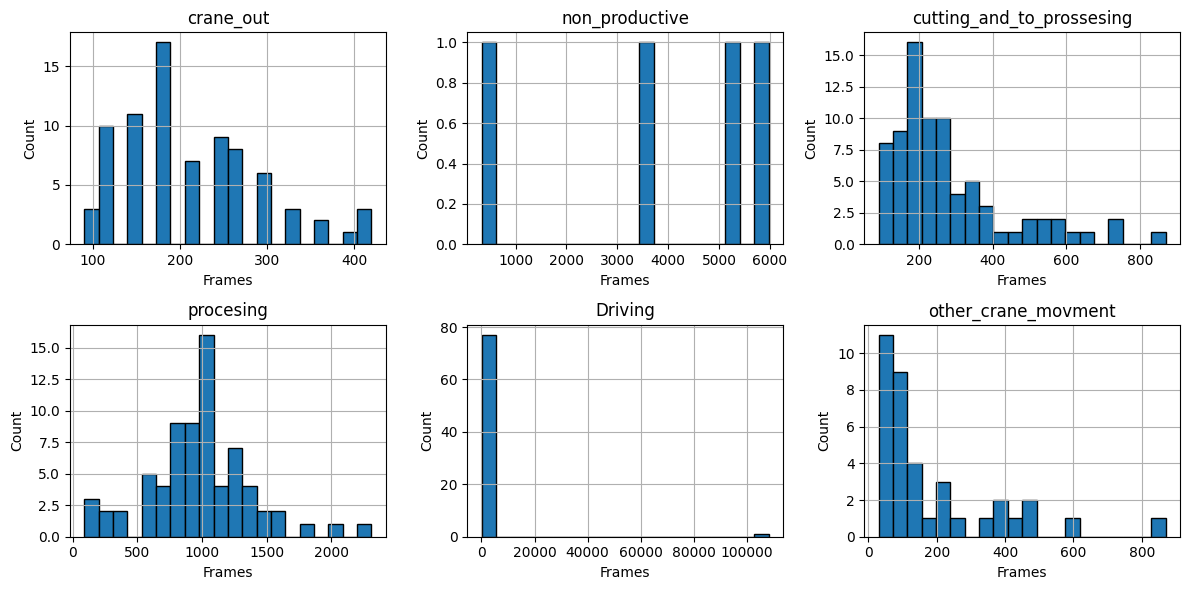}
    \caption{Histograms of frame counts for each class in the dataset. Each subplot shows the distribution of frame lengths for the videos within a specific class.}
    \label{fig:frame_histograms}
\end{figure}

The histograms in Figure~\ref{fig:frame_histograms} illustrate the distribution of frame counts per video clip across each labeled class. Most classes, such as \textbf{crane\_out}, \textbf{cutting\_and\_to\_processing}, and \textbf{processing}, exhibit relatively consistent and compact distributions of clip durations, typically concentrated under 1000 frames. The \textbf{non\_productive} class shows a sparse distribution with a few long-duration clips, reflecting its underrepresentation and irregular occurrence. In contrast, the \textbf{driving} class displays a heavily skewed distribution, where most clips are short, but a few outliers exceed 80,000 frames—likely due to prolonged uninterrupted movement. The \textbf{other\_crane\_movement} class also demonstrates high variability with a right-skewed distribution, indicating inconsistent durations and a broader operational definition. These imbalances and outliers in frame counts may affect model training and class performance, especially for underrepresented or highly variable classes.

\section{Methodology}

The video classification pipeline developed in this project leverages the modular design of PyTorch Lightning \cite{falcon2019pytorch} and the specialized video processing capabilities of PyTorchVideo \cite{pytorchvideo2021}. This integration facilitates efficient training and evaluation of deep learning models tailored for video data.

\subsection{Preprocessing}

Each video clip undergoes uniform temporal subsampling to extract 8 frames, followed by normalization using standard RGB mean and standard deviation values. To enhance model generalization, the following data augmentations are applied during training:

\begin{itemize}
    \item Random short-side scaling (256--320 pixels)
    \item Random cropping to \(244 \times 244\) pixels
    \item Random horizontal flipping
\end{itemize}

For validation, a deterministic preprocessing pipeline is employed, which includes uniform temporal subsampling, normalization, fixed short-side scaling (256 pixels), and center cropping. These preprocessing steps are implemented using PyTorchVideo's transformation utilities \cite{fan2021pytorchvideo}.

\subsection{Model Architecture}

The classification model is based on the 3D ResNet-50 architecture, as implemented in PyTorchVideo \cite{fan2021pytorchvideo}. This architecture extends the traditional 2D ResNet-50 by incorporating spatiotemporal (3D) convolutions, enabling the model to capture both spatial and temporal features from video clips.

\paragraph{Architecture Overview}

The 3D ResNet-50 \cite{hara2018can} comprises the following components:

\begin{itemize}
    \item \textbf{Input Stem}: Applies a 3D convolution with a kernel size of $(1, 7, 7)$, stride of $(1, 2, 2)$, and padding of $(0, 3, 3)$, preserving the temporal dimension while reducing spatial dimensions. This is followed by batch normalization, a ReLU activation function, and a max pooling layer with kernel size $(1, 3, 3)$ and stride $(1, 2, 2)$.
    
    \item \textbf{Residual Stages}: Consists of four residual stages, each containing multiple bottleneck blocks. Each block includes:
    \begin{itemize}
        \item A $1 \times 1 \times 1$ convolution for channel reduction.
        \item A $3 \times 3 \times 3$ convolution for spatiotemporal processing.
        \item A $1 \times 1 \times 1$ convolution for channel restoration.
    \end{itemize}
    Each block utilizes batch normalization and ReLU activation, with skip connections to facilitate gradient flow and model depth. Temporal downsampling is achieved via stride-2 convolutions in the time dimension in selected blocks.
    
    \item \textbf{Head}: After the final residual stage, a global average pooling operation is applied across the temporal and spatial dimensions. The output is passed to a fully connected layer, producing logits for classification. The number of output units corresponds to the number of classes.
\end{itemize}

\paragraph{Adaptations for Video Data}

Key adaptations for processing video data include:

\begin{itemize}
    \item \textbf{3D Convolutions}: All convolutions operate in three dimensions, capturing motion (temporal) and appearance (spatial) features.
    \item \textbf{Flexible Temporal Resolution}: Temporal strides and kernel sizes are adjusted to retain or downsample temporal resolution as needed, based on clip length and computational budget.
    \item \textbf{Batch Normalization and ReLU}: Consistently applied after convolution layers to ensure stable and nonlinear learning dynamics.
\end{itemize}

\paragraph{Implementation Details}

The model is implemented using PyTorchVideo's modular utilities. Configuration details include:

\begin{itemize}
    \item \textbf{Input Channels}: 3 (RGB)
    \item \textbf{Model Depth}: 50 layers (ResNet-50 variant)
    \item \textbf{Number of Classes}: 4 (forestry activity classes)
    \item \textbf{Convolution Type}: 3D convolutions throughout
    \item \textbf{Normalization}: BatchNorm3d
    \item \textbf{Activation}: ReLU
\end{itemize}

\subsection{Training Procedure}

Training is conducted on 4 A100-80GB GPUs, with each clip being 4 seconds in duration. The batch size is set to 8, and training proceeds for up to 200 epochs. The Adam optimizer is employed with a fixed learning rate of 0.001. Mixed precision is disabled, and 32-bit precision is used throughout.
The loss function is cross-entropy. 

The dataset is divided into training and validation sets, focusing on four classes: \texttt{crane\_out}, \texttt{cutting\_and\_to\_processing}, \texttt{Driving}, and \texttt{processing}. Data loaders for both splits are configured with clip-based sampling and multithreading to ensure efficient training. This pipeline ensures reproducibility and scalability, enabling straightforward adaptation for new data or additional classes.

\section{Results}
\subsection{Evaluation Metrics}

To evaluate the performance of the video classification model, several standard classification metrics were computed for both the training and validation phases. These metrics provide a comprehensive understanding of the model’s predictive performance across multiple classes.

\paragraph{Accuracy}
Accuracy measures the proportion of correctly predicted labels out of the total number of predictions:

\[
\text{Accuracy} = \frac{1}{N} \sum_{i=1}^{N} \mathbb{1}(y_i = \hat{y}_i)
\]

where \( y_i \) is the true label, \( \hat{y}_i \) is the predicted label, and \( N \) is the total number of samples.

\noindent
The function \( \mathbb{1}(\cdot) \) is the \textbf{indicator function}, defined as:

\[
\mathbb{1}(A) =
\begin{cases}
1 & \text{if } A \text{ is true} \\
0 & \text{if } A \text{ is false}
\end{cases}
\]

\noindent
It returns 1 if the prediction is correct, and 0 otherwise. The sum therefore counts the number of correct predictions, and dividing by \( N \) yields the accuracy.

\paragraph{Precision (Macro-Averaged)}
Precision indicates how many of the predicted positive instances for each class are actually correct. The macro-averaged precision is computed as:

\[
\text{Precision}_{\text{macro}} = \frac{1}{C} \sum_{c=1}^{C} \frac{\text{TP}_c}{\text{TP}_c + \text{FP}_c}
\]

\paragraph{Recall (Macro-Averaged)}
Recall measures the proportion of actual positive instances that were correctly identified:

\[
\text{Recall}_{\text{macro}} = \frac{1}{C} \sum_{c=1}^{C} \frac{\text{TP}_c}{\text{TP}_c + \text{FN}_c}
\]

\paragraph{F1 Score (Macro-Averaged)}
The F1 score is the harmonic mean of precision and recall, calculated per class and averaged:

\[
\text{F1}_{\text{macro}} = \frac{1}{C} \sum_{c=1}^{C} 2 \cdot \frac{\text{Precision}_c \cdot \text{Recall}_c}{\text{Precision}_c + \text{Recall}_c}
\]

\paragraph{Cross-Entropy Loss}
The loss function used for optimization is the cross-entropy loss, defined as:

\[
\mathcal{L}_{\text{CE}} = - \sum_{i=1}^{N} \sum_{c=1}^{C} y_{i,c} \log(\hat{p}_{i,c})
\]

where \( y_{i,c} \) is the true label (one-hot encoded) and \( \hat{p}_{i,c} \) is the predicted probability for class \( c \).

\vspace{0.5em}
\noindent
These metrics were calculated using the \texttt{torchmetrics} library \cite{torchmetrics2021}  and logged at the end of each epoch using Weights \& Biases (Wandb) to support experiment tracking and performance visualization.

\subsection{Experimental Results}
\begin{table}[H]
\centering
\caption{Final evaluation metrics after epoch 190 of training using validation dataset}
\label{tab:final_results}
\begin{tabular}{|l|c|}
\hline
\textbf{Metric} & \textbf{Value} \\
\hline
Training F1 Score         & 0.96 \\
Validation F1 Score       & 0.88 \\
Validation Precision      & 0.90 \\
Validation Recall         & 0.88 \\
\hline
\end{tabular}
\end{table}

\begin{figure}
    \centering
    \includegraphics[width=0.7\linewidth]{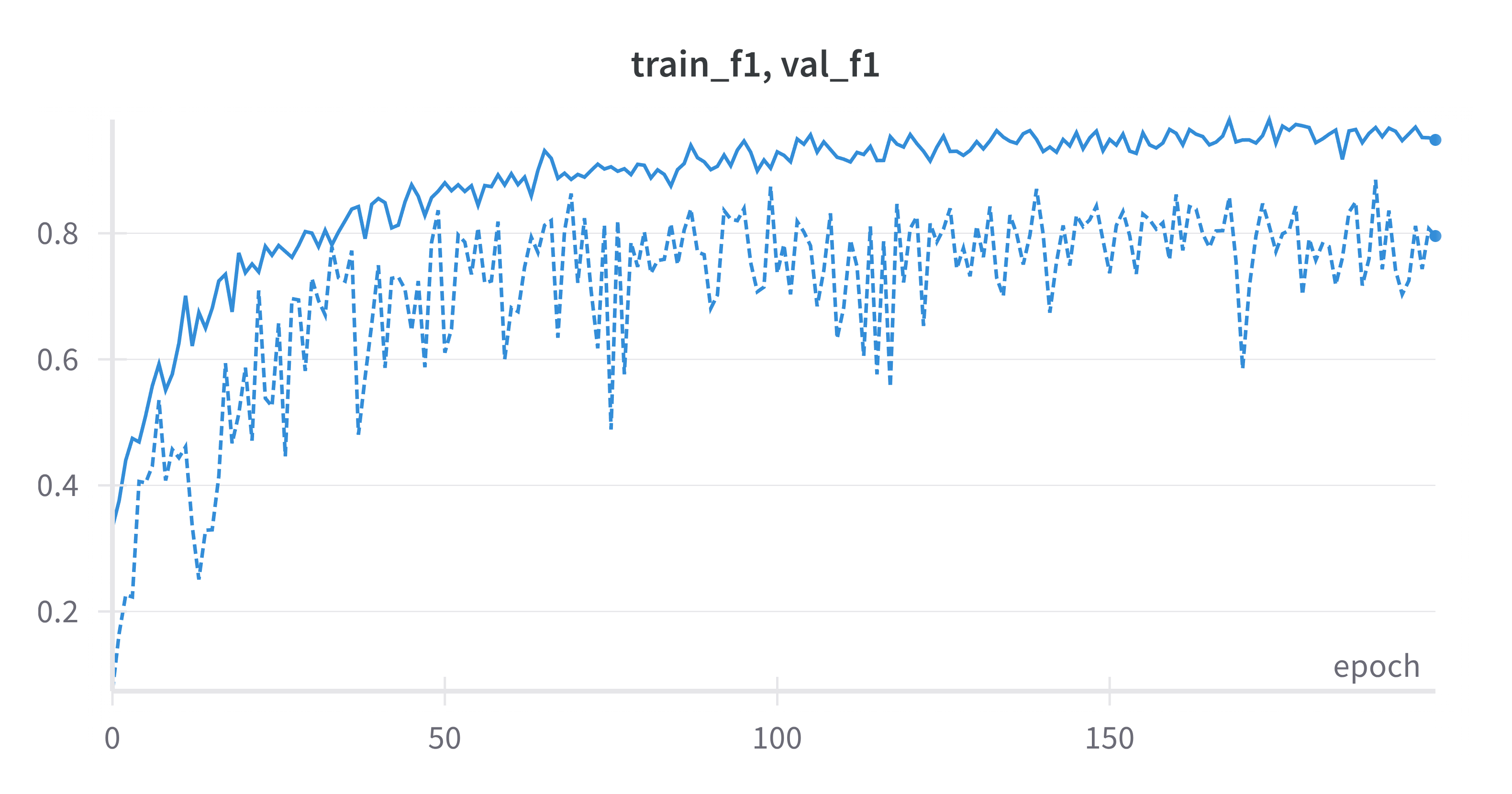}
    \caption{Train and Val f1 with a visible overfitting resulting from small dataset. Dotted line denots validation.}
    \label{fig:train_val_f1}
\end{figure}

\section{Conclusions and future work}

The experimental results demonstrate that the proposed 3D ResNet-50-based classification pipeline is capable of learning meaningful spatiotemporal representations for recognizing forestry work elements from dashcam footage. A validation F1 score of 0.88 and precision of 0.90 indicate that the model can generalize well across several operational classes, especially in a constrained four-class setup. These results are promising for a pilot-scale dataset and suggest the potential for automating time studies in forest operations.

However, several limitations must be addressed before this approach can be scaled for broader deployment. The most significant issue observed during training was overfitting. While the training F1 score reached 0.96, the divergence from the validation F1 score points to the model's limited generalization, likely due to the relatively small dataset size. This is further supported by the gap between training and validation loss values. The model likely memorized patterns from the training data, which restricts its ability to handle unseen conditions or edge cases in forestry operations.

Another challenge involves class imbalance, particularly for the \texttt{non\_productive} and \\ \texttt{other\_crane\_movement} categories, which were excluded from model training due to their low representation. This imbalance affects the scalability of the model to more fine-grained or rare work phases. Future efforts should consider targeted data collection or synthetic oversampling techniques (e.g., SMOTE or data augmentation) to address underrepresented categories.

The model architecture itself showed strong capacity to encode both spatial and temporal features. However, more advanced techniques such as attention mechanisms, transformers, or hybrid CNN-RNN models could be explored to improve temporal reasoning and reduce reliance on fixed-length input clips. Additionally, integrating sensor data (e.g., crane angle, GPS, machine status) could provide multi-modal context to complement the video stream.

Finally, although the current pipeline is designed for offline analysis, adapting the system for real-time inference on embedded platforms remains an important future milestone. Optimizations such as quantization, model pruning, or distillation could be applied to reduce the inference time and resource requirements, enabling deployment in on-board computing environments.

In summary, the pilot results validate the feasibility of deep learning-based video classification in forestry, but also reveal the need for larger datasets, broader class coverage, and more efficient model variants for real-world deployment.

\section{Code and data}
Training code available at \url{https://gitlab.nibio.no/maciekwielgosz/forest_video_classification}

Data will be provided here soon.


\bibliographystyle{plain}
\bibliography{references}

\end{document}